\documentclass[10pt,twocolumn,letterpaper]{article}

\usepackage{cvpr}              

\definecolor{cvprblue}{rgb}{0.21,0.49,0.74}
\usepackage[pagebackref,breaklinks,colorlinks,allcolors=cvprblue]{hyperref}

\def\paperID{10160} 
\def\confName{CVPR}
\def\confYear{2025}

\title{The Devil is in Low-Level Features for Cross-Domain Few-Shot Segmentation}

\author{Yuhan Liu,\quad Yixiong Zou\thanks{Corresponding author.},\quad Yuhua Li,\quad  Ruixuan Li\\
	School of Computer Science and Technology, Huazhong University of Science and Technology\\
	{\tt\small \{yuhan\_liu, yixiongz, idcliyuhua, rxli\}@hust.edu.cn}
}

\usepackage{amsmath}
\usepackage{amsfonts}
\usepackage{newtxtext} 
\usepackage{bm}
\usepackage{graphicx}
\usepackage{multirow}

\begin{document}
\maketitle

\begin{abstract}
Cross-Domain Few-Shot Segmentation (CDFSS) is proposed to transfer the pixel-level segmentation capabilities learned from large-scale source-domain datasets to downstream target-domain datasets, with only a few annotated images per class. 
In this paper, we focus on a well-observed but under-explored phenomenon in CDFSS: for target domains, particularly those distant from the source domain, segmentation performance peaks at the very early epochs, and declines sharply as the source-domain training proceeds.
We delve into this phenomenon for an interpretation: low-level features are vulnerable to domain shifts, leading to sharper loss landscapes during the source-domain training, which is the devil of CDFSS. 
Based on this phenomenon and interpretation, we further propose a method that includes two plug-and-play modules: one to flatten the loss landscapes for low-level features during source-domain training as a novel sharpness-aware minimization method, and the other to directly supplement target-domain information to the model during target-domain testing by low-level-based calibration. 
Extensive experiments on four target datasets validate our rationale and demonstrate that our method surpasses the state-of-the-art method in CDFSS signifcantly by 3.71\% and 5.34\% average MIoU in 1-shot and 5-shot scenarios, respectively.
\end{abstract}

\section{Introduction}

Current deep neural networks~\cite{long2015fully,ronneberger2015u,zhao2017pyramid} have achieved remarkable success in semantic segmentation, but their performance heavily relies on large-scale annotated datasets~\cite{benenson2019large,lin2014microsoft}. However, the process of annotating data, particularly for dense pixel-wise tasks like semantic segmentation, is labor-intensive and time-consuming. To address this issue, Few-Shot Segmentation (FSS) has been introduced, aiming to generate pixel-level predictions for unseen categories with only a few labeled samples. Typically, the model is pretrained on a large-scale dataset of base classes and then transferred to unseen novel classes. However, novel classes might not be in the same domain as base classes~\cite{everingham2010pascal,li2020fss,demir2018deepglobe,codella2019skin,tschandl2018ham10000}, a more realistic approach is to consider domain gaps between the base (source domain) and novel (target domain) classes, which gives rise to the Cross-Domain Few-Shot Segmentation (CDFSS) task~\cite{lei2022cross}.
\begin{figure}[t]
	\centering
	\includegraphics[width=1.0\columnwidth]{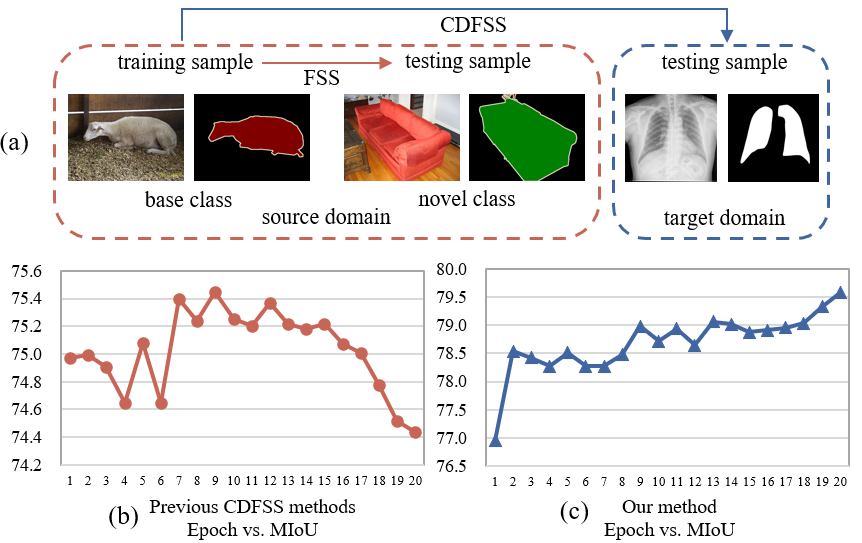} \vspace{-0.65cm}
	\caption{(a) In CDFSS tasks, the training (source) and testing (target) datasets belong to distinct domains, with categories in the testing dataset being unseen during training. (b) Previous CDFSS methods show a decreasing trend of mIoU as the source-domain training proceeds, even at very early epochs for distant domains. (c) Our method can effectively prevent the model from performance decline after early epochs and achieve higher performance.}\vspace{-0.5cm}
	\label{fig:intro1}
\end{figure}

Although extensive works~\cite{lei2022cross,herzog2024adapt,su2024domain,he2024apseg} have been developed, a well-observed phenomenon is still not handled: for target domains, especially those distant from the source domain, the best performance is always achieved at the very early epochs or even the first epoch. As shown in Fig.~\ref{fig:intro1}b, the performance decreases sharply as the source-domain training proceeds, where the 20th epoch's MIoU is even lower than that of the first epoch. Although the early stop can handle this problem, the goal of the source-domain training is to provide a generalizable model for all target domains~\cite{lei2022cross}. Therefore, it may be inappropriate to try different early stops for specific domains.
In this paper, we aim to delve into this well-observed but under-explored phenomenon and handle it based on our interpretations of it, as shown in Fig.~\ref{fig:intro1}c.

\begin{figure}
	\centering
	\includegraphics[width=1.0\columnwidth]{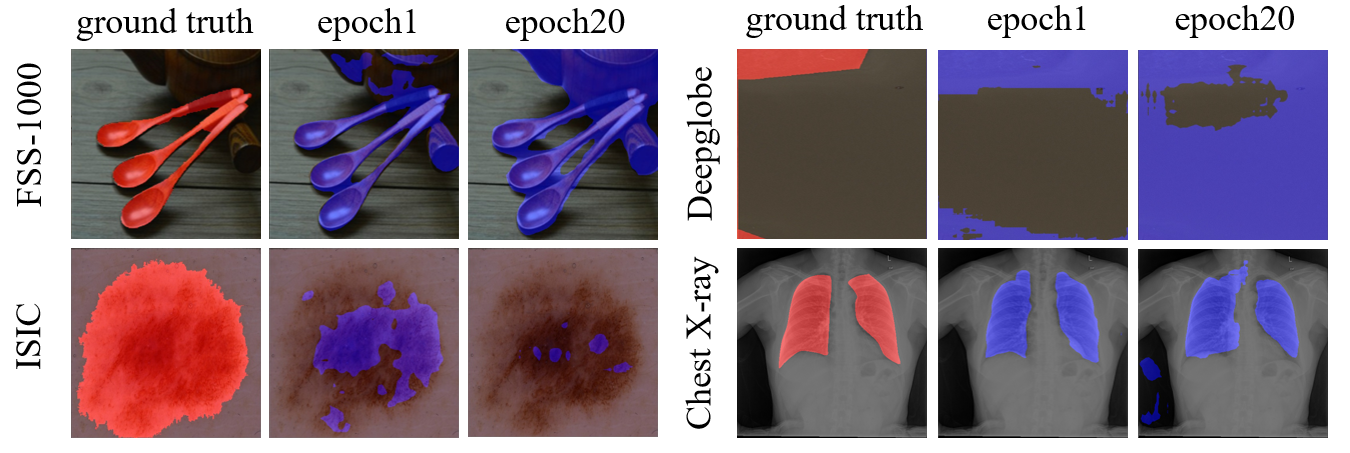} \vspace{-0.6cm}
	\caption{The visualization of predictions at the 1st and 20th epoch further indicate that CDFSS models may not acquire meaningful information for target domains during source-domain training.}\vspace{-0.5cm}
	\label{fig:intro2}
\end{figure}

To delve into this phenomenon, we first visualize the predictions of the 1st and 20th epochs in Fig.~\ref{fig:intro2}. We observe that the 20th epoch's model shows many fundamental errors in target domains, i.e., CDFSS models fail to identify the foreground but focus on entirely wrong regions. This suggests that the CDFSS models may not have acquired meaningful information for target domains during source-domain training, resulting in almost random predictions during testing. 
We then visualize feature maps at different network stages, and observe this problem occurs even at the shallow layers, and deeper layers produce wrong feature maps with wrong inputs from shallow layers.
Quantitatively, 
previous works~\cite{foret2020sharpness,zou2024flatten} suggest that the challenge posed by domain shift can be analyzed through the flatness of the loss landscape. We use this as an entry point to explore the connection between shallow layers and early stops.
We find that low-level features, although simpler and previously believed to be more transferable~\cite{zou2021revisiting}, are vulnerable to domain shifts, leading to sharper loss landscapes during the source-domain training. That is, the devil lies in the low-level features.

Based on this phenomenon and interpretation, we further propose a method for the CDFSS task.  
During the source-domain training, we propose a novel sharpness-aware minimization method to flatten the loss landscapes for low-level features, which is achieved by a shape-preserving perturbation module with random convolutions.
During the target-domain phase, since low-level features can hardly capture the target-domain information due to its vulnerability to domain shifts, we directly supplement such information to the model with a low-level-based calibration module, which effectively prevents the model from fundamental errors caused by collapsed low-level features.
Based on our methods, we can effectively prevent the model from dropping performances after early epochs and achieve higher performance (Fig.~\ref{fig:intro1}c).

In summary, our contributions can be listed as 

$\bullet$ We focus on a well-observed but under-explored phenomenon in CDFSS: for target domains, particularly those distant from the source domain, the best segmentation performance is always achieved at the very early epochs, followed by a sharp decline as source-domain training progresses.

$\bullet$ We delve into this phenomenon for an interpretation: low-level features are vulnerable to domain shifts, leading to sharper loss landscapes during the source-domain training, which is the devil of CDFSS. 

$\bullet$ Building on this interpretation, we propose a method that includes two plug-and-play modules: one act as a novel sharpness-aware minimization method to flatten the loss landscapes of low-level features during source-domain training, while the other directly supplement target-domain information to the model during target-domain testing.

$\bullet$ Extensive experiments on four target datasets validate the rationale of our interpretation and method, showing its superiority over current state-of-the-art methods.

\section{Interpretation}

\subsection{Preliminaries}
Cross-Domain Few-Shot Segmentation (CDFSS) aims to transfer the segmentation capabilities learned from the source domain to target domains, with only a few annotated images per class. The source domain $\mathcal{D}_s=(\mathcal{X}_s, \mathcal{Y}_s)$ and target domain $\mathcal{D}_t=(\mathcal{X}_t, \mathcal{Y}_t)$ are defined by different input distributions and disjoint label spaces, i.e., $\mathcal{X}_s\neq\mathcal{X}_t$, $\mathcal{Y}_{s}\cap \mathcal{Y}_{t}=\emptyset $, where $\mathcal{X}$ refers to input distribution and $\mathcal{Y}$ refers to the label space.

In this work, we adopt the meta-learning episodic paradigm following~\cite{lei2022cross}. Specifically, both $\mathcal{D}_s$ and $\mathcal{D}_t$ consist of several episodes. Each episode is constructed by a support set $S = {\{(I_s^i, M_s^i)\}}_{i=1}^K$ with $K$ training samples and a query set $Q = \{(I_q, M_q)\}$, where $I$ denote the image and $M$ denote the pixel labels. Within each episode, the model leverages $\{I_s, M_s\}$ and $I_q$ to predict the query mask $M_q$.

The model is trained on the source domain support sets and query set $\left \{ S^s,Q^s \right \}$  by minimizing the binary cross-entropy loss:
\begin{equation}
		\mathcal{L}=BCE \left ( F\left ( I_s^s,I_q^s,M_s^s \right ) ,M_q^s  \right ),
        \label{eq: BL_loss}
\end{equation}
where $F\left ( I_s^s,I_q^s,M_s^s \right )=h\left ( f\left ( I_s^s,I_q^s \right ) ,M_s^s \right )$ outputs the segmentation score map of $I_q^s$, which is composed of a feature extractor $f\left(\cdot \right)$ and a comparison module $h\left(\cdot \right)$.

Then the model will be evaluated on the target domain query set $Q^t$, the prediction for $I_q^t$ is:
\begin{equation}
		\hat{M_q^t}=arg\max F\left ( I_s^t,I_q^t,M_s^t \right )  
        \label{eq: predict}
\end{equation}
Finally, the evaluation can be conducted based on the comparison between $\hat{M_q^t}$ and the real label $M_q^t$.

\subsection{Delve into the Early Stop in Cross-Domain Few-Shot Segmentation}

\subsubsection{Intuitive observation: The devil is in the low-level features}

\begin{figure}
	\centering
	\includegraphics[width=0.85\columnwidth]{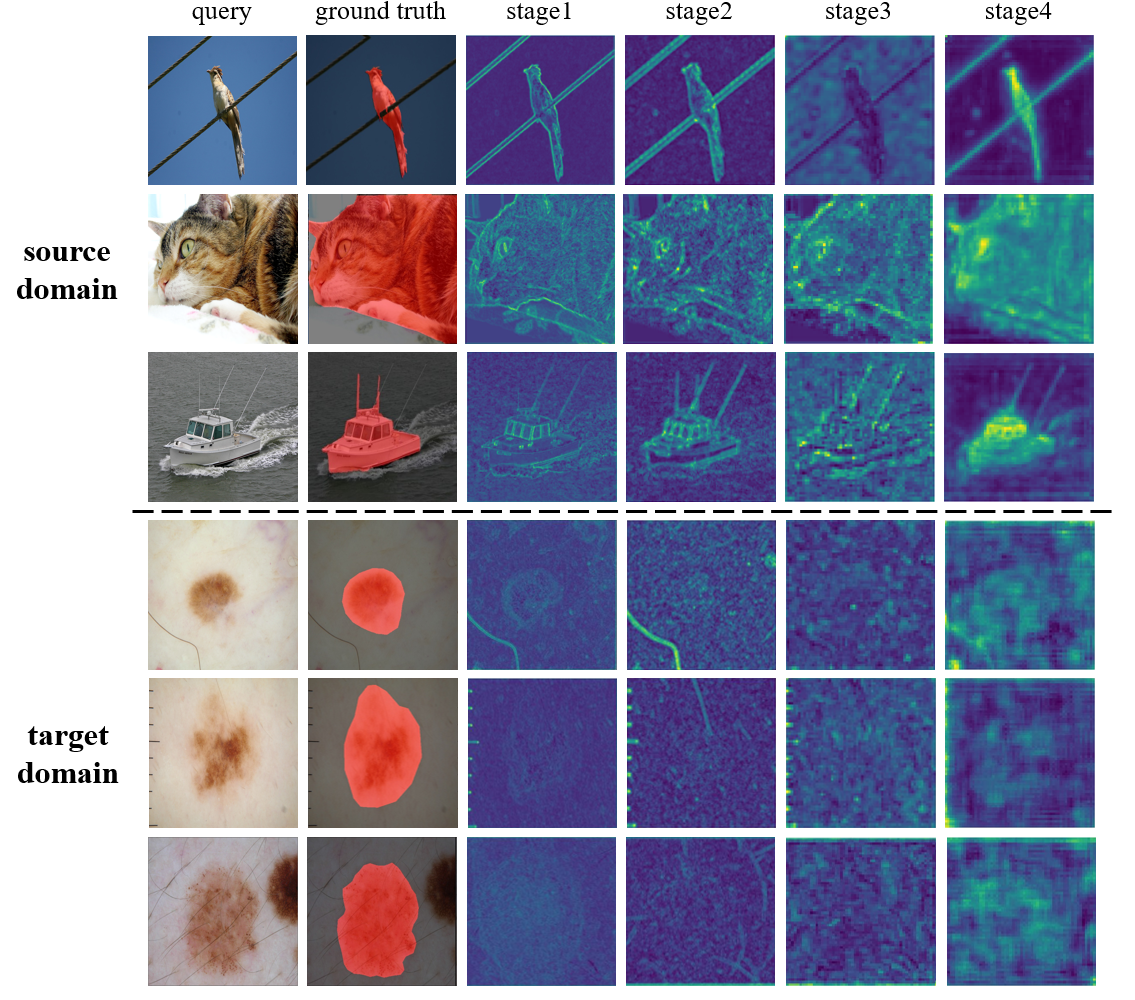} 
        \vspace{-0.2cm}
	\caption{Feature maps from stages 1 to 4 for both the source domain and target domains. The noticeable contrast in the feature maps at stage 1 between source and target domains indicates the limited performance of CDFSS models stems from shallow layers.}\vspace{-0.2cm}
	\label{fig:low_level_feature_vis1}
\end{figure}

Given our suspicion that the model fails to learn useful information from the source domain for target domains, we start by visualizing the feature maps. Using ResNet-50~\cite{he2016deep} as the backbone following~\cite{fan2022self}, we visualize the feature maps of stages 1 to 4 for both the source domain and the target domains, as shown in Fig.~\ref{fig:low_level_feature_vis1}. We observe that on source-domain samples, the model learns boundary information effectively in stage 1, enabling it to accurately focus on the foreground in stage 4. However, on target domains especially those distant from the source domain, even though the foreground is clear and prominent, the model still fails to learn anything meaningful in stages 1 and 2. Consequently, in stage 4, the model focuses on completely incorrect areas. This indicates that the model's bad performance may originate from its shallow layers.

\begin{figure}
	\centering
	\includegraphics[width=1.0\columnwidth]{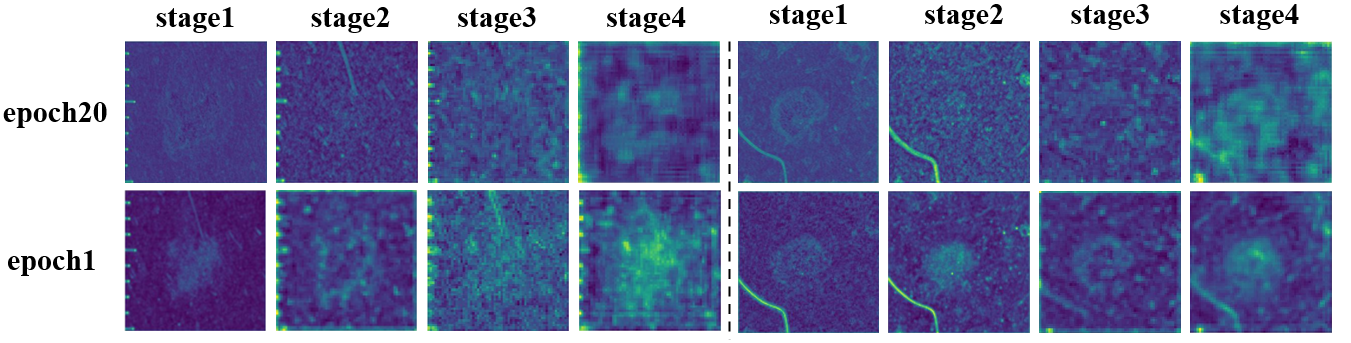} 
        \vspace{-0.6cm}
	\caption{Feature maps from stages 1 to 4 for epoch 1 and epoch 20. Epoch 1 in stage 1 shows more distinguishable activations than epoch 20, indicating that low-level features gradually incorporate incorrect information as training progresses.}\vspace{-0.55cm}
	\label{fig:low_level_feature_vis2}
\end{figure}

In Fig.\ref{fig:low_level_feature_vis2}, we further visualize feature maps of each stage at the 1st and 20th epoch, respectively. We find that the shallow layers show more distinguishable activations at epoch 1 than at epoch 20, which aligns with our observation of higher accuracy at epoch 1. Consequently, we conclude that the poor performance of CDFSS models may be attributed to shallow layers that fail to capture useful low-level information, a trend that intensifies during source-domain training.

\subsubsection{Quantitative verification: Low-level features lead to a sharper loss landscape}

\begin{figure}
	\centering
	\includegraphics[width=1.0\columnwidth] {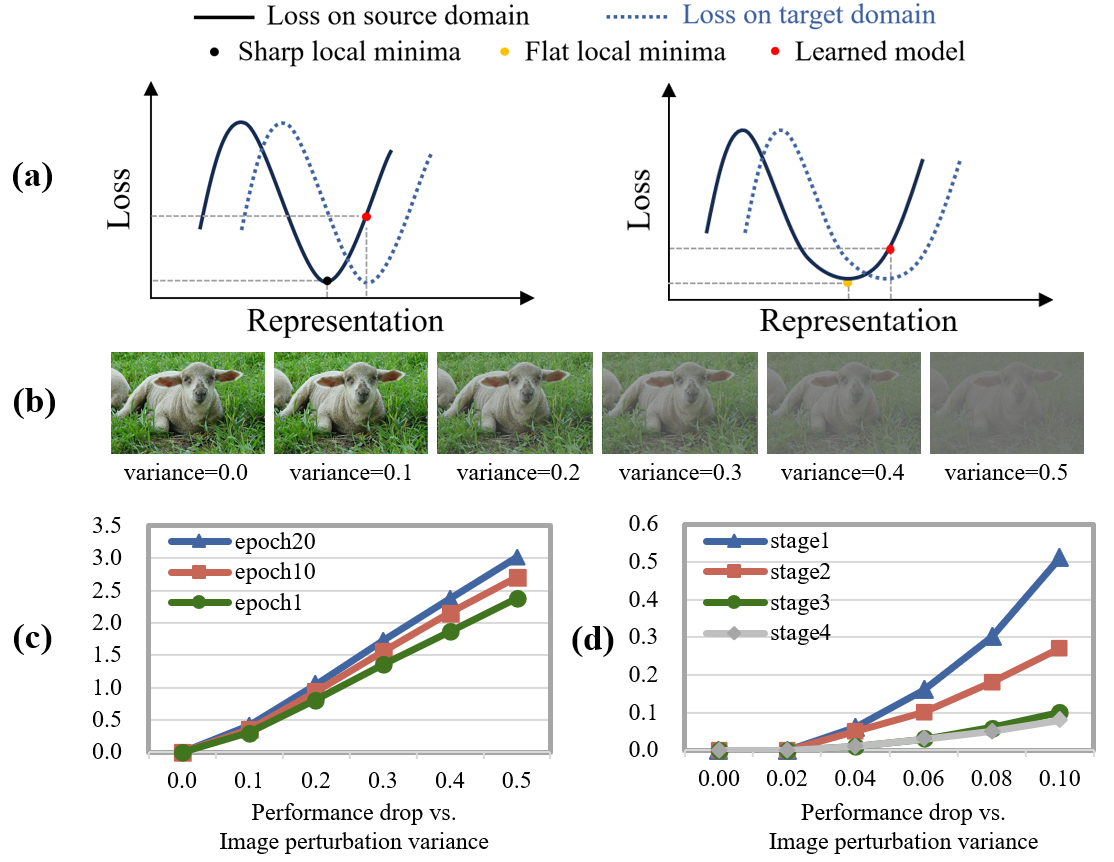} \vspace{-0.5cm}
	\caption{(a) A sharp minimum in the landscape corresponds to a representation that is highly sensitive to data shifts. (b) Examples of pixel perturbation applied to the images. (c) As training proceeds, the loss landscape becomes progressively sharper. (d) Perturbing shallow layers leads to much sharper loss landscapes, indicating shallow layers are the cause of the sharp loss landscape and the increased sensitivity to domain shifts.}\vspace{-0.5cm}
	\label{fig:loss_landscape_epoch_layer}
\end{figure}

Previous work~\cite{foret2020sharpness,zou2024flatten} has demonstrated that the sharpness of the loss landscape can serve as a tool to analyze cross-domain issues. As illustrated in Fig.~\ref{fig:loss_landscape_epoch_layer}a, a smoother loss landscape enables the model to better handle domain shifts\footnote{Please refer to the appendix for more details about the sharpness.}. 
Therefore, we measure the sharpness of the source-domain-trained models against different epochs and stages.

We first measure the sharpness of the loss landscape across different epochs. Following the method in ~\cite{zou2024flatten}, we apply pixel perturbations to the training data. Examples of these perturbations are shown in Fig.~\ref{fig:loss_landscape_epoch_layer}b, while the magnitude of performance drop is depicted in Fig.\ref{fig:loss_landscape_epoch_layer}c. A larger performance drop under the same perturbation indicates a sharper loss landscape. As observed, the performance of models across all epochs drops when exposed to perturbations, but the model from epoch 1 experiences a smaller drop than that of epoch 10, which in turn performs better than epoch 20. This suggests that as training progresses, the loss landscape becomes progressively sharper, making the model increasingly vulnerable to domain shifts and resulting in decreased performance on target domains. This is in line with our previous observations.

Based on this, we further investigate the relationship between model layers and the loss landscape. We apply the same magnitude of perturbations to the feature maps at stages 1-4 and measure the performance drops. The results in Fig.\ref{fig:loss_landscape_epoch_layer}d show that earlier stages exhibit larger performance drops under the same perturbation, which implies that shallow layers lead to sharper loss landscapes and are more vulnerable to domain shifts. This experiment provides quantitative validation for the intuitive observation from the low-level feature visualization in the previous section.

\subsubsection{Further analysis: Low-level features absorb domain-specific information}
\begin{table}
	\belowrulesep=0pt
	\aboverulesep=0pt
	\renewcommand\arraystretch{1.1}
	\centering
	\caption{The model with fixed shallow layers performs better than the one with trained shallow layers, implying that shallow layers learn domain-specific information that negatively impacts generalization to the target domain.} \vspace{-0.2cm}
	\label{tab:train_or_fix_stage}
	\resizebox{1.0\columnwidth}{!}{
		\begin{tabular}{c|cccc|c}
			\toprule
			Method & FSS-1000 & Deepglobe & ISIC  & ChestX-ray & Average \\
			\midrule
			train stage1, 2, 3, 4 & 78.86 & 39.44 & 35.76 & 72.49 & 56.64 \\
			fix stage1, train stage2, 3, 4 & 78.88 & 39.90 & 37.00 & 72.12 & 56.98 \\
			fix stage1, 2, train stage3, 4 & 78.91 & 40.00 & 35.49 & 74.44 & 57.21 \\
			\bottomrule
	\end{tabular}}%
\vspace{-0.2cm}
\end{table}%

\begin{figure}
	\centering
	\includegraphics[width=1.0\columnwidth]{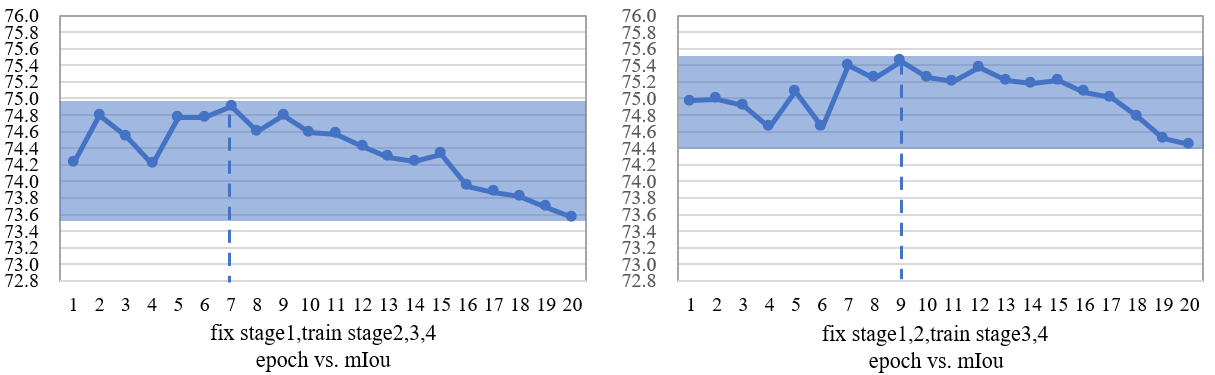}\vspace{-0.2cm}
	\caption{The mIoU curves over epochs for two different training settings indicate that shallow layers are more vulnerable to domain shifts, leading to more severe performance degradation during source-domain training.}\vspace{-0.4cm}
	\label{fig:fixstage_miou}
\end{figure}

To investigate the impact of shallow layers on model performance, we compare the results of fixing stage 1 and stage 2 using ImageNet~\cite{deng2009imagenet} pre-trained weights with the results of training all stages, as shown in Tab.~\ref{tab:train_or_fix_stage}. We find that fixing stage 1 and stage 2 leads to the best performance. 

In Fig~\ref{fig:fixstage_miou}, we further visualize the mIoU curves over epochs for two different training settings: fix stage 1 only, and fix stage 1, 2. The comparison revealed that when stage 2 is fixed, the model reaches the optimal mIoU at a later epoch, and as training progresses, the mIoU decreases more slowly. 

This suggests that shallow layers are vulnerable to domain shifts, therefore any slight absorption of domain information could harm target-domain generalization. 

\subsection{Conclusion and Discussion}
Based on the above experiments, we make the interpretation as follows. The shallow layers of the model tend to learn domain-specific information, causing the low-level features to become increasingly domain-specific during training, gradually overfitting to the source domain. The overfitting sharpens the model's loss landscape and makes it more vulnerable to domain shifts, thereby reducing cross-domain performance and introducing significant mistakes in CDFSS.
This problem is partially addressed by the early-stop strategy and freezing shallow layers.

However, merely stopping the training of shallow layers cannot fully unlock the power of source-domain training. Therefore, below we propose a novel method, \textbf{LoEC} (\textbf{Lo}w-level \textbf{E}nhancement and \textbf{C}alibration), to handle the problem in shallow layers.

\section{Method}
\begin{figure*}
	\centering
	\includegraphics[width=0.78\linewidth]{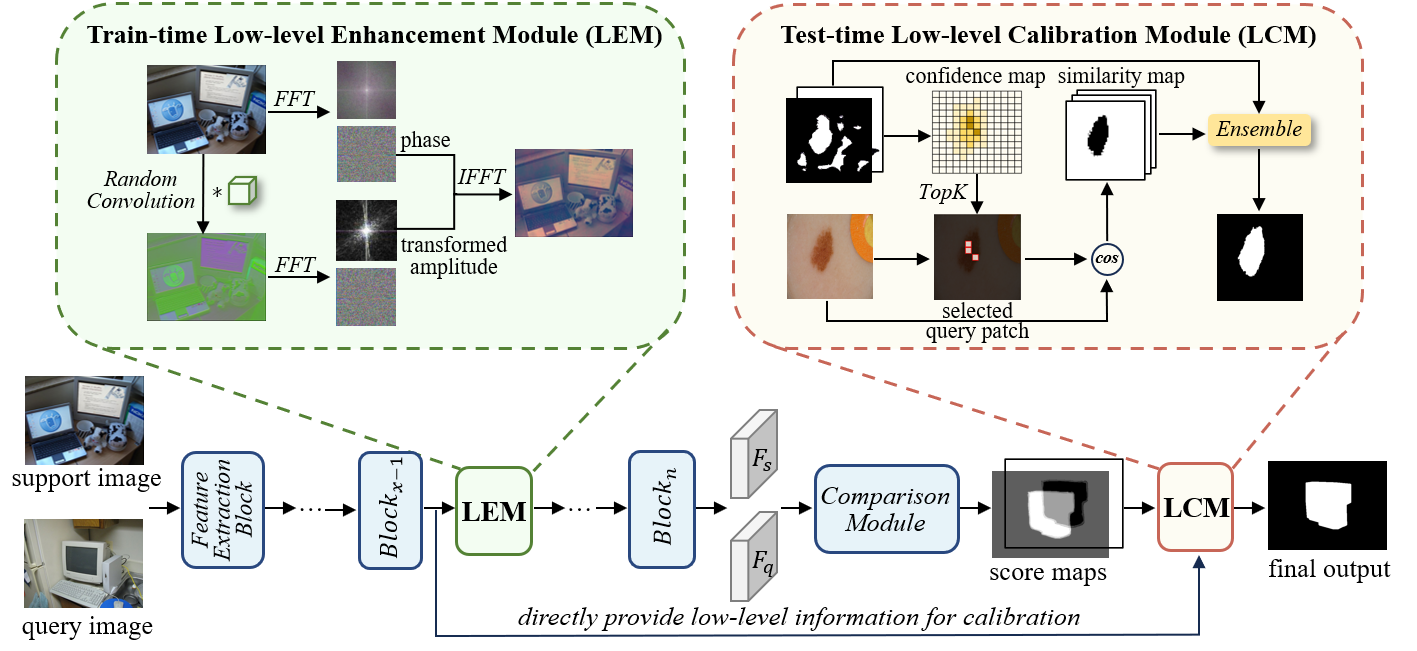} \vspace{-0.1cm}
	\caption{The overall architecture of our method in a 1-way 1-shot example. Our method consists of two modules: the Low-level Enhancement Module (LEM) and the Low-level Calibration Module (LCM). Support and query images are fed into a weight-shared backbone to extract features, and then processed by a comparison module to obtain a coarse segmentation mask. During source-domain training, LEM perturbs low-level support features for sharpness-aware minimization. During target-domain testing, LCM calibrates the coarse segmentation mask by directly supplementing collapsed low-level target-domain information.\protect\footnotemark} \vspace{-0.5cm}
	\label{fig:method}
\end{figure*}
\footnotetext{The low-level support and query features in LEM and LCM are illustrated as images for better understanding.}

\subsection{Method Overview}
Our core idea is that low-level features are vulnerable to domain shifts, which may lead to a sharp decrease in generalization with only slight absorption of source-domain information. Based on this, we designed two plug-and-play modules to operate in two phases: 1) During the source-domain training phase, the Low-level Enhancement Module enhances the robustness against domain shifts by synthesizing diverse domains and incorporating them into low-level features. 2) During the target-domain testing phase, the Low-level Calibration Module refines previous segmentation results by utilizing the low-level features from the target domain. The framework is illustrated in Fig.~\ref{fig:method}.

\subsection{Train-time Low-level Enhancement Module}
As analyzed above, overfitting of low-level features on the source domain during training leads to a sharp loss landscape, making the model vulnerable to domain shifts. According to the previous Sharpness Aware Minimization (SAM) method \cite{foret2020sharpness,zou2024flatten}, one effective strategy to smooth loss landscapes is to introduce perturbations. Thus we transform the domain of the low-level support feature during training, which acts as a form of perturbation to reduce overfitting.

Previous work \cite{xu2021robust} has shown that random convolution is a shape-preserving strategy to distort local textures, which we view to be suitable for the segmentation task. We take advantage of this property to generate diverse domains while retaining the content. Specifically, given support feature $F_s\in \mathbb{R}^{H\times W \times C} $ from the early layers of the encoder and a convolution layer with filters $\Theta \in \mathbb{R}^{h \times w \times C \times C}$, where $H$, $W$ and $C$ represent height, width and channels of the input feature, and $h$ and $w$ are the height and width of the random convolution filter. We sample the filter weights from $N(0, \sigma^2)$, where $\sigma$ acts as a hyperparameter to control the perturbation magnitude. Then we get perturbed support feature $F_s^{\prime}$ with transformed domain:
{\vspace{-0.25cm}
	\begin{equation} 
		F_s^{\prime} = F_s * \Theta,
\vspace{-0.25cm} \end{equation}}
Although random convolution can generally preserve the shape and semantics of the feature, some edge details inevitably get blurred. Existing research \cite{oppenheim1981importance} has shown that the Fast Fourier Transformation (FFT) decomposes a signal into amplitude spectrum and phase spectrum, with amplitude capturing the overall domain and texture information and phase representing shape, edge, and spatial structure. Thus, we apply FFT to both $F_s$ and $F_s^{\prime}$, decomposing them into phase spectrum $\mathcal{P}$, $\mathcal{P^{\prime}}$ and amplitude spectrum $\mathcal{A}$, $\mathcal{A^{\prime}}$:
{\vspace{-0.2cm}
	\begin{equation}
		\mathcal{A} e^{i \mathcal{P}}=F F T(F_s),
\vspace{-0.4cm} \end{equation}}
{\vspace{-0.2cm}
	\begin{equation}
		\mathcal{A^{\prime}} e^{i \mathcal{P^{\prime}}}=F F T(F_{s}^{\prime}),
\vspace{-0.1cm} \end{equation}}

Then we recombine the phase from the original feature $F_s$ and the amplitude from the perturbed feature $F_s^{\prime}$, using the Inverse Fast Fourier Transform (IFFT) to generate transformed support feature:
{\vspace{-0.2cm}
	\begin{equation}
		F_s^{t}=I F F T\left(\mathcal{A}^{\prime} e^{i \mathcal{P}}\right),
\vspace{-0.2cm} \end{equation}}

Therefore, while transforming the domain, we achieve better preservation of the boundary shapes, which achieves domain-oriented perturbations on low-level features for sharpness-aware minimization. The transformed feature is then fed into the subsequent layers of the encoder. 

\subsection{Test-time Low-level Calibration Module}
After being processed by the encoder, the support and query features are fed into the comparison module, yielding a coarse score map $S \in \mathbb{R}^{H\times W \times 2} $. 
Since low-level features may be collapsed on target domains due to the vulnerability to domain shifts, we directly use the low-level query feature $F_{q} \in \mathbb{R}^{H^{\prime} \times W^{\prime} \times C}$ to supplement such collapsed low-level target-domain information to calibrate the score map during testing on the target domain.

First, we compute a confidence map $C$ by subtracting the background similarity from the foreground similarity in the score map:
{
	\begin{equation}
	C_{i, j}=S_{i, j, 1}-S_{i, j, 0},
\end{equation}}

Here, $S_{i, j, 1}$ and $S_{i, j, 0}$ denote the foreground and background similarity for the pixel at $\left(i,j\right)$, respectively. The confidence map $C$ quantifies the likelihood of foreground at each pixel.

Then we partition the confidence map into patches $\left \{ P_1, P_2,\dots,P_n  \right \} $ and select the Top-K patches with the highest average confidence. These patches are considered the most reliable foreground regions:
{\vspace{-0.2cm}
	\begin{equation}
		\bar{C}_p = \frac{1}{|P_p|} \sum_{(i, j) \in P_p} C(i, j) ,
 \vspace{-0.2cm} \end{equation}}
{\vspace{-0.15cm}
	\begin{equation}
        \{ P_{k_1}, P_{k_2}, \dots, P_{k_K} \} = \arg \max_{P_p} \bar{C}_p
 \vspace{-0.2cm} \end{equation}}

Where $\left | P_{p}  \right | $ is the number of pixels in patch $P_{p}$ and $\left(i,j\right)$ are the pixel locations within the patch.

For each of the selected Top-K foreground patches, we locate the corresponding patch in the interpolated query feature map $F_{q}^{\prime}\in \mathbb{R}^{H \times W \times C}$ as $\{ F_{k_1}, F_{k_2}, \dots, F_{k_K} \} $, and compute the cosine similarity between it and all other patches in the query feature map:
{\vspace{-0.2cm}
	\begin{equation}
		Sim(F_{k_i},F_m) = \frac{F_{k_i} \cdot F_m}{\|F_{k_i}\| \|F_m\|}
 \vspace{-0.15cm} \end{equation}}

Finally, the foreground similarity in the score map is updated using the top-K similarity maps:
{\vspace{-0.15cm}
	\begin{equation} 
        S_{i,j,1}^{\prime} = S_{i,j,1} + \sum_{k=1}^{K} w_k \cdot \left( Sim_k\left ( i,j \right ) - \beta_k \right),
 \vspace{-0.15cm} \end{equation}}

Here, $w_k$ is a scaling factor and $\beta_k$ is a bias term. This adjustment improves foreground detection by directly supplementing collapsed low-level target-domain information.

During the source-domain training phase, the model will be trained by minimizing the loss in Eq.~\ref{eq: BL_loss}.
During the target-domain testing phase, the prediction will be obtained based on Eq.~\ref{eq: predict} and LCM.

\section{Experiments}

\subsection{Datasets}
Following the settings in~\cite{lei2022cross}, we use PASCAL VOC 2012~\cite{everingham2010pascal} with SBD augmentation~\cite{hariharan2011semantic} as source domain for training. Then we evaluate the trained models on four target domains: FSS-1000~\cite{li2020fss}, Deepglobe~\cite{demir2018deepglobe}, ISIC~\cite{codella2019skin,tschandl2018ham10000}, and Chest X-ray~\cite{candemir2013lung,jaeger2013automatic}. See Appendix for more details.

\subsection{Implementation Details}
We employ ResNet-50~\cite{he2016deep} with ImageNet~\cite{deng2009imagenet} pretrained weights as our backbone. Following baseline SSP~\cite{fan2022self}, we discard the last backbone stage and the last ReLU for better generalization. Furthermore, we integrated our approach into the transformer architecture by using ViT-B/16~\cite{dosovitskiy2020image} as the backbone, following FPTrans~\cite{zhang2022feature}. Consistent with FPTrans~\cite{zhang2022feature}, we resize both support and query images to $480\times480$. During training on the source domain, we use SGD to optimize our model, with a momentum of 0.9 and a initial learning rate of 1e-3. We apply our LEM into different shallow layers of the backbone, more details can be seen in \ref{sec:Sensitivity Study of Hyper-parameters}. The random convolution filter used in LEM has a kernel size of $3\times3$, and the standard deviation $\sigma$ is set to be 0.1. During testing on the target domain, we apply our LCM to refine segmentation mask.
The hyperparameters $K$, $w$ and $\beta$ are set to 3, 0.6 and 0.7, respectively.

\subsection{Comparison with State-of-the-art Methods}

\begin{table*}[htbp]
	\belowrulesep=0pt
	\aboverulesep=0pt
	\renewcommand\arraystretch{1.2}
	\centering
	\caption{MIoU of 1-shot and 5-shot results compared with previous FSS and CD-FSS methods. All the methods are trained on PASCAL and tested on the CD-FSS. The best performance among all methods is highlighted in bold.} \vspace{-0.2cm}
	\label{tab:sota}
	\resizebox{0.88\textwidth}{!}{
		\begin{tabular}{c|c|c|cc|cc|cc|cc|cc}
			\toprule[1pt]
			\multirow{2}[2]{*}{Method} & \multirow{2}[2]{*}{Mark} & \multirow{2}[2]{*}{Backbone} & \multicolumn{2}{c|}{FSS-1000} & \multicolumn{2}{c|}{Deepglobe} & \multicolumn{2}{c|}{ISIC} & \multicolumn{2}{c|}{Chest X-ray} & \multicolumn{2}{c}{Average} \\
			\cmidrule{4-13}          &       &       & \multicolumn{1}{c|}{1-shot} & 5-shot & \multicolumn{1}{c|}{1-shot} & 5-shot & \multicolumn{1}{c|}{1-shot} & 5-shot & \multicolumn{1}{c|}{1-shot} & 5-shot & \multicolumn{1}{c|}{1-shot} & 5-shot \\
			\midrule
			\multicolumn{13}{c}{Few-Shot Segmentation Methods} \\
			\midrule
			RePRI~\cite{boudiaf2021few}  & CVPR-21 & Res-50  & 70.96 & 74.23 & 25.03 & 27.41 & 23.27 & 26.23 & 65.08 & 65.48 & 46.09 & 48.34 \\
			HSNet~\cite{min2021hypercorrelation} & ICCV-21  & Res-50  & 77.53 & 80.99 & 29.65 & 35.08 & 31.20 & 35.10 & 51.88 & 54.36 & 47.57 & 51.38 \\
			SSP~\cite{fan2022self}   & ECCV-22  & Res-50  & 78.91 & 80.59 & 40.00 & 48.68 & 35.49 & 45.86 & 74.44 & 74.26 & 57.21 & 62.35 \\
			FPTrans~\cite{zhang2022feature} & NIPS-22 & ViT-base  & 80.74 & 83.65 & 38.36 & 49.30 & 48.65 & 60.37 & 80.92 & 82.91 & 62.17 & 69.06 \\
			PerSAM~\cite{zhang2024personalize}  & ICLR-24 & ViT-base  & 60.92 & 66.53 & 36.08 & 40.65 & 23.27 & 25.33 & 29.95 & 30.05 & 37.56 & 40.64 \\
			\midrule
			\multicolumn{13}{c}{Cross-Domain Few-Shot Segmentation Methods} \\
			\midrule
			PATNet~\cite{lei2022cross}  & ECCV-22  & Res-50  & 78.59 & 81.23 & 37.89 & 42.97 & 41.16 & 53.58 & 66.61 & 70.20 & 56.06 & 61.99 \\
                APM~\cite{tonglightweight} & NIPS-24 & Res-50 & 79.29 & 81.83 & 40.86 & 44.92 & 41.71 & 51.16 & 78.25 & 82.81 & 60.03 & 65.18 \\
			ABCDFSS~\cite{herzog2024adapt}& CVPR-24  & Res-50  & 74.60 & 76.20 & 42.60 & 49.00 & 45.70 & 53.30 & 79.80 & 81.40 & 60.67 & 64.97 \\
			DRA~\cite{su2024domain}   & CVPR-24  & Res-50  & 79.05 & 80.40 & 41.29 & 50.12 & 40.77 & 48.87 & 82.35 & 82.31 & 60.86 & 65.42 \\
			APSeg~\cite{he2024apseg} & CVPR-24  & ViT-base  & 79.71 & 81.90 & 35.94 & 39.98 & 45.43 & 53.98 & \pmb{84.10} & \pmb{84.50} & 61.30 & 65.09 \\
			\textbf{LoEC} & Ours  & Res-50  & 78.51 & 80.60 & \pmb{44.10} & 49.67 & 38.21 & 47.04 & 81.02 & 82.73 & 60.46 & 65.01 \\
			\textbf{LoEC} & Ours  & ViT-base  & \pmb{81.05} & \pmb{83.69} & 42.12 &\pmb{51.48}& \pmb{52.91} &\pmb{62.43}& 83.94 & 84.12 & \pmb{65.01} & \pmb{70.43}\\
			\bottomrule[1pt] 
		\end{tabular}%
		\label{tab:addlabel}%
	} \vspace{-0.2cm}
\end{table*}%

\begin{figure}
	\centering
	\includegraphics[width=0.6\columnwidth]{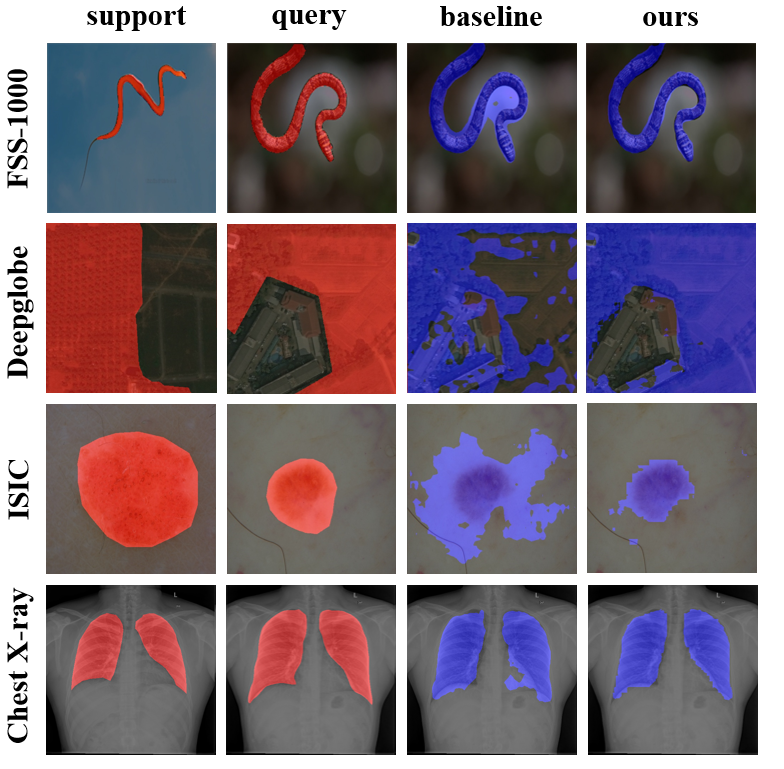}\vspace{-0.2cm}
	\caption{Qualitative results of our model for 1-shot setting.}
	\label{fig:qualitative_results} \vspace{-0.4cm}
\end{figure}

In Tab.~\ref{tab:sota}, we compare our method with CNN-based and ViT-based approaches, including traditional FSS methods and existing CDFSS methods. Our results illustrate a notable enhancement for both 1-shot and 5-shot tasks. Specifically, our method outperforms the state-of-the-art APSeg~\cite{he2024apseg} by 3.71\% and 5.34\% in the 1-shot and 5-shot settings respectively, confirming the effectiveness of our strategy. Additionally, we apply our method on existing CDFSS baselines and observe improved performance. Please refer to the appendix for more details. We present qualitative results of our method in 1-way 1-shot segmentation in Fig.\ref{fig:qualitative_results}.

\subsection{Ablation Study}
\subsubsection{Effectiveness of Each Component}
\begin{table}
	\centering
	\caption{Ablation study on key components.} \vspace{-0.2cm}
	\resizebox{0.5\columnwidth}{!}{
		\begin{tabular}{cccc}
            \toprule
            LEM   & LCM   & ResNet & ViT \\
            \midrule
                  &       & 57.21 & 62.17 \\
            $\checkmark$     &       & 58.35 & 63.06 \\
                  & $\checkmark$     & 59.78 & 64.39 \\
            $\checkmark$     & $\checkmark$     & \textbf{60.46} & \textbf{65.01} \\
            \bottomrule
            \end{tabular}%
		\label{tab:component}%
	}\vspace{-0.2cm}
\end{table}%

We evaluate each proposed component on both CNN and ViT baselines to assess the effectiveness of our designs, including LEM and LCM. The results presented in Tab.~\ref{tab:component} show that in the 1-shot setting, introducing LEM improved average MIoU by 1.14\% for the CNN baseline and 0.89\% for the ViT baseline, while adding LCM further increased it by 2.11\% and 1.95\%, respectively. These results clearly demonstrate the effectiveness of each component in enhancing performance.

\subsubsection{Verification of LEM}
\textbf{LEM smooths loss landscapes.}\footnote{We present the validation of CNN-based methods here, and similar trends hold for ViT-based methods. Please refer to the appendix for details.}
\begin{figure}
	\centering
	\includegraphics[width=1.0\columnwidth]{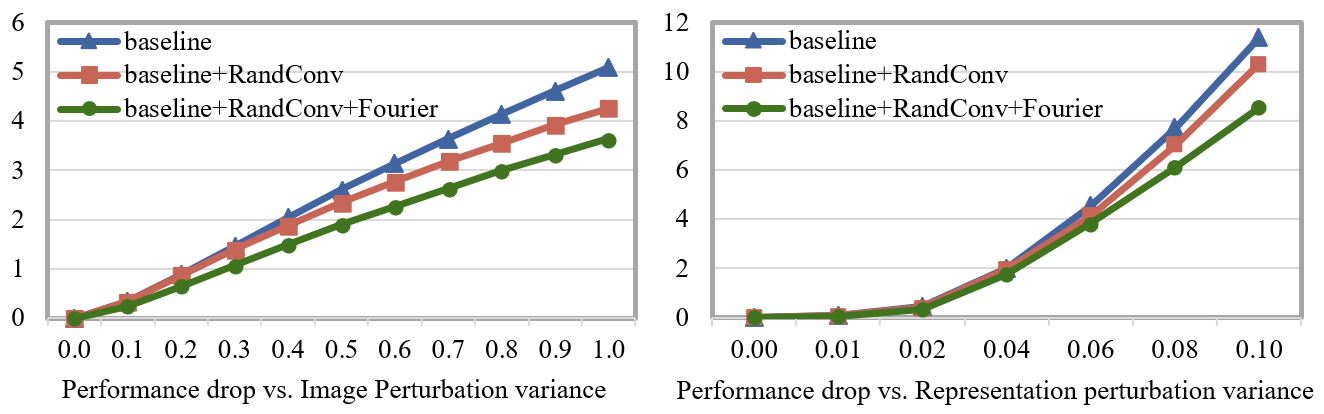} \vspace{-0.5cm}
        \caption{Applying LEM leads to reduced loss landscape sharpness and improved robustness to domain shifts, Fourier transformation further enhances these benefits.}
	\label{fig:LEM_loss_landscape}\vspace{-0.3cm}
\end{figure}
To assess the impact of LEM, we measured the loss landscape sharpness before and after its application. We introduced low-frequency noise on training data to simulate domain shifts in the representation space (i.e., pixels and features). A larger performance drop suggests a sharper landscapes. As illustrated in Fig.\ref{fig:LEM_loss_landscape}, random convolution smooths the loss landscape compared with the baseline, while Fourier transformation further enhances this effect. These results confirm that LEM effectively smooths the loss landscape, which improves robustness against domain shifts.

\noindent\textbf{LEM ensures performance improves with training.}
\begin{figure}
	\centering
	\includegraphics[width=1.0\columnwidth]{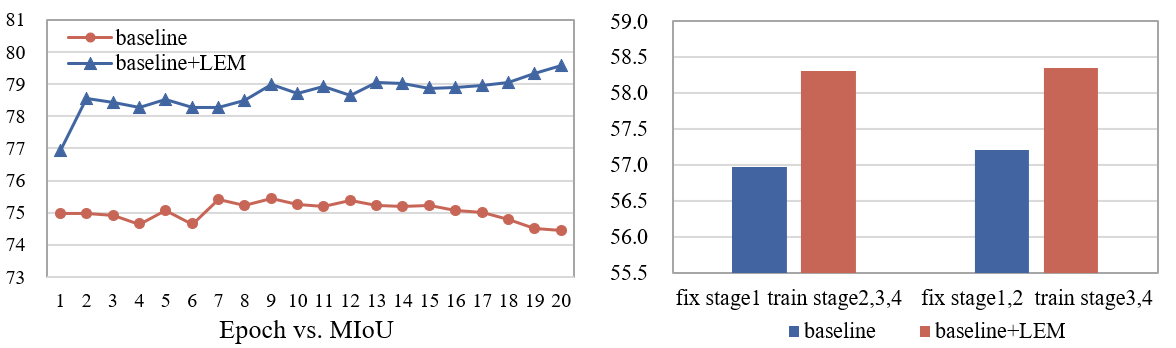} \vspace{-0.5cm}
	\caption{(Left) The mIoU of Chest-X Ray with and without LEM, confirms that LEM leads to consistent and steady performance improvement as training advances. (Right) After incorporating our LEM module, training shallow layers can achieve performance comparable to fixing them.}\vspace{-0.5cm}
	\label{fig:LEM_epoch_fixstage}
\end{figure}
We compared mIoU trends over training epochs with and without LEM in Fig.\ref{fig:LEM_epoch_fixstage} (left). In the baseline model, mIoU on the target domain declines as it increasingly absorbs source-domain information during training. Conversely, with LEM, mIoU not only outperforms the baseline at each epoch but also keeps increasing as training advances, indicating that the model focuses more on domain-invariant information.

\noindent\textbf{LEM prevents the shallow layers from absorbing excessive source-domain information.}
As previously analyzed, fixing stage 1 and stage 2 of the baseline resulted in the best performance, which suggests that shallow layers tend to learn information that negatively impacts target-domain generalization. Fig.\ref{fig:LEM_epoch_fixstage} (right) shows that after incorporating our LEM, training stage 2 produced results comparable to those achieved by fixing stage 2. This indicates that LEM effectively prevents the shallow layers from absorbing excessive source-domain information during the training process.

\noindent\textbf{LEM encourages the model to focus more on domain-agnostic information.}
\begin{figure}
	\centering
	\includegraphics[width=1.0\columnwidth]{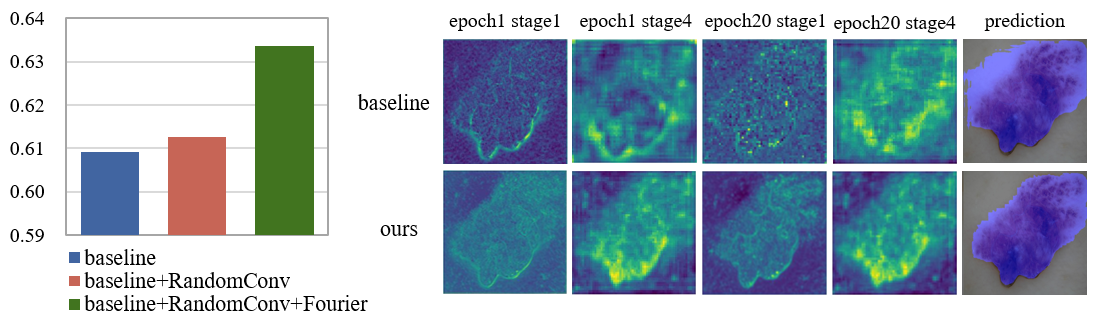} \vspace{-0.5cm}
	\caption{(Left) The CKA similarity between source and target domain with and without LEM, verifies that LEM helps the model to focus more on domain-agnostic information. (Right) Feature map visualization of stage 1 and stage 4 at epoch 1 and epoch 20.}\vspace{-0.5cm}
	\label{fig:stage_LEM}
\end{figure}
To assess how the domain distance between the source and target domains changes before and after applying LEM, we use CKA (Centered Kernel Alignment) similarity for measurement, following the approach in \cite{davari2022reliability,zou2024attention,zou2024closer,zou2024compositional,zou2022margin}. Specifically, we extract features from source domain and target domain images using a backbone network, then compute the CKA similarity of the final layer's feature by aligning the channel dimension.  A higher CKA similarity indicates a smaller domain distance, suggesting the model retains less domain-specific information. As shown in Fig.\ref{fig:stage_LEM} (left), the average CKA similarity across four target domains increases after applying random convolution, and is further improved with the addition of the Fourier transform. This demonstrates that by perturbing the source domain, LEM encourages the model to capture domain-agnostic information during training.

\noindent\textbf{LEM enables the model to capture more useful low-level information in shallow layers.}
As shown in Fig.\ref{fig:stage_LEM} (right), LEM helps the model learn boundary information more effectively in stage 1, allowing it to focus more accurately on the foreground in stage 4. This leads to more precise activation regions in the later epochs of source domain training.

\subsubsection{Verification of LCM}
\textbf{LCM smooths loss landscapes.}
\begin{figure}[t]
	\centering
	\includegraphics[width=1.0\columnwidth]{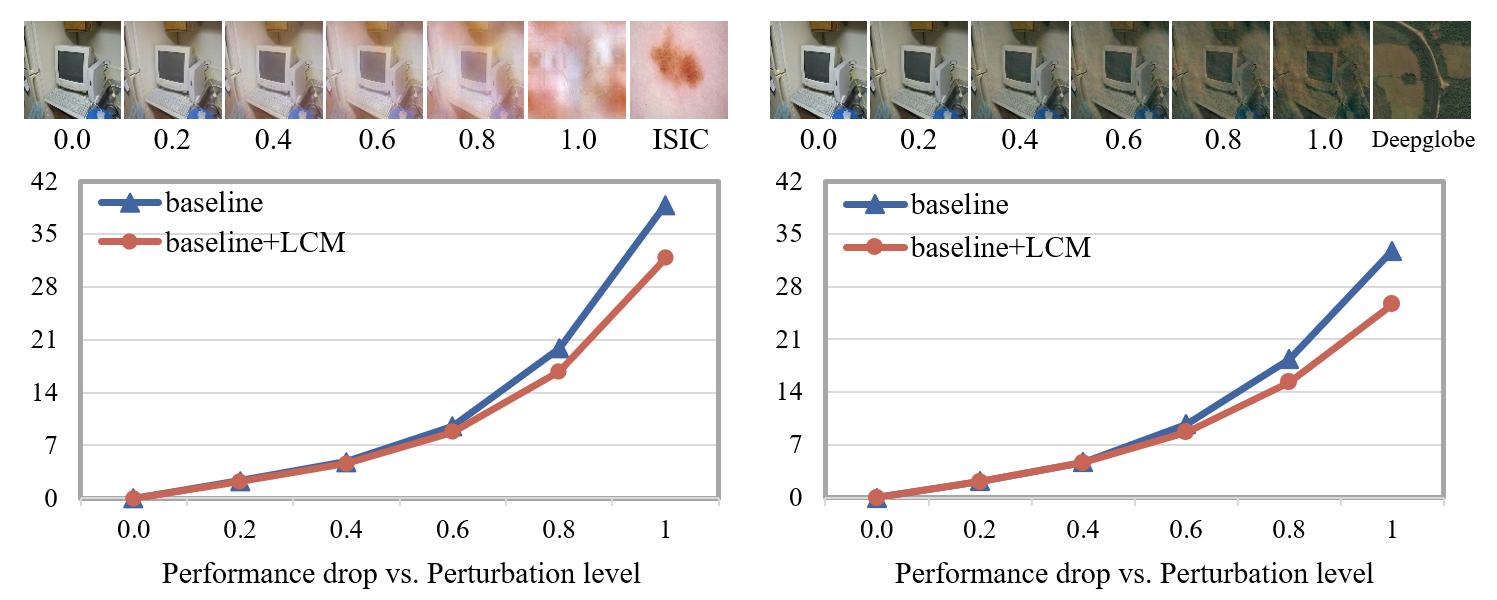} \vspace{-0.5cm}
	\caption{LCM reduces the sharpness of the loss landscape and enhances robustness against domain shifts.} \vspace{-0.5cm}
	\label{fig:LCM_loss_landscape}
\end{figure}
To assess the impact of LCM in reducing sharpness and enhancing robustness against domain shifts, we superimpose styles from ISIC medical images and Deepglobe remote sensing images onto source domain images to simulate domain shifts, with the level of style alteration serving as the perturbation level. As shown in Fig.\ref{fig:LCM_loss_landscape}, the lowered performance drop validates that LCM smooths the loss landscape, thus improves robustness against domain shifts.

\subsubsection{Sensitivity Study of Hyper-parameters}
\label{sec:Sensitivity Study of Hyper-parameters}
\begin{figure}[t]
	\centering
	\includegraphics[width=1.0\columnwidth]{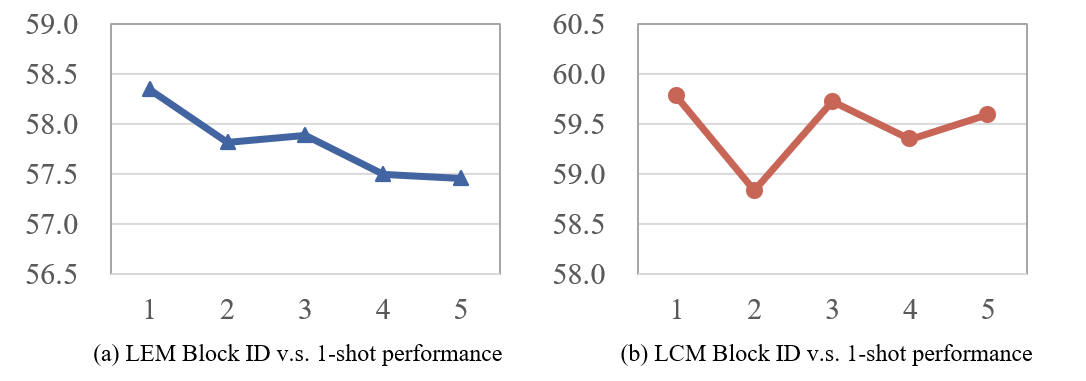} 
        \vspace{-0.6cm}
	\caption{The effect of LEM and LCM at different positions in the shallow layers. Block ID represents the sequence in which the module is applied: before stage 1, stage 1 block 1, stage 2 block 1, stage 2 block 2, and stage 2 block 3.}\vspace{-0.3cm}
	\label{fig:Hyper-parameters} \vspace{-0.2cm}
\end{figure}
As shown in Fig.\ref{fig:Hyper-parameters}, we investigated the effects of using LEM and LCM at various positions in the shallow layers. Please refer to the appendix for more details.

\section{Related Work}

\subsection{Few-Shot Segmentation}
Few-Shot Segmentation (FSS) aims to segment novel classes in query images using only a few annotated support samples. Existing methods in FSS can be generally categorized into two types: prototype-based methods and matching-based methods. Motivated by PrototypicalNet\cite{snell2017prototypical}, prototype-based methods \cite{lang2022learning,li2021adaptive,tian2020prior,wang2019panet} generate prototypes from support images and perform segmentation by computing the similarity between query features and these support prototypes. On the other hand, matching-based methods\cite{lu2021simpler,min2021hypercorrelation,peng2023hierarchical,zhang2021few} focus on analyzing pixel-to-pixel dense correspondences between support and query features to avoid the loss of spatial structure inherent in prototype-based methods. However, these methods are limited to segmenting novel classes within a single domain and do not generalize effectively to unseen domains.
\subsection{Domain Generalization}
Domain Generalization (DG) aims to train models that can generalize to diverse, unseen target domains, particularly when target domain data is unavailable during training, which aligns with the objective of CDFSS. Existing methods for domain generalization can be categorized into two types: methods that focus on learning domain-invariant feature representations across multiple source domains\cite{carlucci2019domain,wang2019learning}, and methods that generate diverse samples through data or feature augmentation\cite{du2020learning,shankar2018generalizing}. However, most existing DG methods focus on generalizing the whole model, which is not suitable for direct use in CDFSS tasks. In contrast, our plug-and-play LEM module is compact, lightweight, and effective. In the appendix, we compare our LEM module with domain generalization methods to demonstrate the effectiveness of our approach as a novel sharpness-aware minimization method in addressing the CDFSS problem.

\subsection{Cross-domain Few-Shot Segmentation}
Unlike traditional FSS setting, CDFSS requires models to generalize to unseen target domains without accessing target data during training, making the task more realistic and complex. PATNet \cite{lei2022cross} introduces a transformation module that converts domain-specific features into domain-agnostic features, which can be finetuned with target domain data during the testing phase. ABCDFSS \cite{herzog2024adapt} proposes a test-time task adaptation method, utilizing consistency-based contrastive learning to prevent overfitting and improve class discriminability. DR-Adapter \cite{su2024domain} introduces a small adapter for rectifying diverse target domain styles to the source domain. Compared with them, our approach proposes two simple plug-and-play modules that require no training and can be directly applied to boost performance in CDFSS. Besides, a few works \cite{fan2022self, nie2024cross} are based on the self-support concept. Our LCM module follows a similar principle but offers two significant advantages: (1) While other methods use query prototypes to match query features, we directly use the low-level query features to adjust the query mask, preventing further loss of information. (2) LCM can be conveniently added during the target-domain testing phase, rather than during the entire training process.

\section{Conclusion}

In this paper, We focus on a well-observed but under-explored phenomenon in CDFSS: for target domains, the best segmentation performance is always achieved at the very early epochs, followed by a sharp decline as source-domain training progresses.
We delve into this phenomenon for an interpretation, and find low-level features are vulnerable to domain shifts, leading to sharper loss landscapes during the source-domain training, which is the devil of CDFSS. 
Based on the interpretation, we further propose a method that includes two plug-and-play modules: one to flatten the loss landscapes for low-level features during source-domain training, and the other to directly supplement target-domain information to the model during target-domain testing.
Extensive experiments on four CDFSS benchmarks validate our rationale and effectiveness. 

\section*{Acknowledgments}

This work is supported by the National Key Research and Development Program of China under grant 2024YFC3307900; the National Natural Science Foundation of China under grants 62206102, 62436003, 62376103 and 62302184 ; Major Science and Technology Project of Hubei Province under grant 2024BAA008; Hubei Science and Technology Talent Service Project under grant 2024DJC078; and Ant Group through CCF-Ant Research Fund. The computation is completed in the HPC Platform of Huazhong University of Science and Technology.

{
    \small
    \bibliographystyle{ieeenat_fullname}
    \bibliography{main}
}

\end{document}